\documentclass[11pt]{article}

\usepackage[final]{acl}

\usepackage{times}
\usepackage{latexsym}
\usepackage{algorithm}
\usepackage{algpseudocode}
\usepackage{booktabs}
\usepackage{float} % enables [H] :contentReference[oaicite:2]{index=2}
\usepackage{amssymb}
 \usepackage{amsmath}
\usepackage{tcolorbox}
\usepackage{lipsum}
\usepackage{adjustbox}
\usepackage{listings}
\usepackage{fvextra}
\usepackage[T1]{fontenc}
\usepackage[utf8]{inputenc}

\usepackage{microtype}

\usepackage{inconsolata}

\usepackage{graphicx}

\usepackage{booktabs}
\usepackage{xcolor}
\usepackage{colortbl}
\usepackage{array}

\definecolor{selected}{RGB}{255, 243, 205}
\definecolor{colA}{RGB}{192, 0, 0}     
\definecolor{colB}{RGB}{46, 117, 182}   
\definecolor{colC}{RGB}{56, 142, 60}  
\definecolor{oursgray}{RGB}{220,220,220}

\colorlet{wrong}{red!25}
\colorlet{selected}{yellow!40}
\colorlet{distractor}{gray!20}

\title{ITLC at SemEval-2026 Task 11: Normalization and Deterministic Parsing for Formal Reasoning in LLMs}

\author {
    Wicaksono Leksono Muhamad$^{*1, 2}$, Joanito Agili Lopo$^{*1, 2}$, Tack Hwa Wong$^{*1, 3}$, \\
    \textbf{Muhammad Ravi Shulthan Habibi$^{1,4}$, Samuel Cahyawijaya\thanks{Equal contributions.}$^{1,5}$} \\
    $^1$SEACrowd$\quad$$^2$Mantera Studio$\quad^3$Universiti Teknologi PETRONAS\quad \\
    $^4$Universitas Indonesia\quad$^5$Cohere \\
    \texttt{\{wcksnlxn,amalopo99,tackhwawong00\}@gmail.com} \\ 
    \texttt{muhammadravi251001@gmail.com, samuelcahyawijaya@cohere.com} \\ 
    Code: \url{https://github.com/SEACrowd/ITLC_semeval2026_shared_task_11} \\
    \\
}

\begin{document}
\maketitle
\begin{abstract}

% Large language models exhibit significant content effects in reasoning tasks, limiting their robustness and reliability across multilingual settings. This work introduces a novel method that reduces these biases through explicit structural abstraction, transforming syllogistic arguments into canonical representations that preserve only logical structure, followed by deterministic parsing to determine validity. Evaluated on the SemEval-2026 Task 11 multilingual benchmark, our approach achieves top-5 rankings across all four subtasks, demonstrating strong validity accuracy while substantially mitigating content effects, offering a competitive and interpretable alternative to complex fine-tuning or activation-level interventions.

Large language models suffer from content effects in reasoning tasks, particularly in multilingual contexts. We introduce a novel method  that reduces these biases through explicit structural abstraction that transforms syllogisms into canonical logical representations and applies deterministic parsing to determine validity. Evaluated on the SemEval-2026 Task 11 multilingual benchmark, our approach achieves top-6 rankings across all subtasks while substantially reducing content effects and offering a competitive alternative to complex fine-tuning or activation-level interventions.

\end{abstract}

\section{Introduction}
The scope to which large language models (LLMs) can perform content-independent reasoning remains a central question in reasoning tasks. Prior work has shown that LLMs exhibit strong \emph{content effects} in real-world knowledge and belief during pre-training \citep{dasgupta2024languagemodelshumanlikecontent, bertolazzi-etal-2024-systematic}. These findings raise concerns about robustness, bias, and reliability in LLM applications.

Recent works have explored different mitigation methods to this problem. For instance, \citet{kim-etal-2025-reasoning} show that LLMs develop specific inference mechanisms in their internal architecture. Similarly, \citet{valentino2025mitigatingcontenteffectsreasoning} introduce kNN-based conditional steering in the architecture to reduce content effect. Meanwhile, Neuro-symbolic and quasi-symbolic approaches have also been explored to improve faithfulness and logical consistency \citep{ranaldi-etal-2025-improving,quan-etal-2024-verification,xu-etal-2024-faithful,lyu-etal-2023-faithful}. Despite these advances, there is still no simple and effective solution for disentangling content from formal reasoning, particularly in multilingual settings.

In this work, we introduce a novel unbiased method for syllogistic reasoning that reduces content effects through explicit structural abstraction. Our approach transforms each argument into a canonical syllogistic representation that preserves only its logical structure, followed by deterministic structural parsing to determine validity. This simple strategy substantially reduces content effects while achieving strong validity accuracy. 

We evaluate our method on the SemEval-2026 Task 11~\cite{valentino-etal-2026-semeval}, a multilingual benchmark for syllogistic reasoning that explicitly measures both validity accuracy and the magnitude of content effects. Our method ranks in the top-5 across 3 subtasks among all participants and 6th position in subtask 2 (See Appendix \ref{sec:leaderboard_comparison} for detail leaderboard comparison). These results demonstrate that our structural abstraction approach remains a competitive and interpretable alternative to heavy fine-tuning~\cite{ranaldi-etal-2025-improving,bertolazzi-etal-2024-systematic} or latent-level interventions~\cite{valentino2025mitigatingcontenteffectsreasoning,lopo-etal-2025-language} for mitigating reasoning biases in both English and multilingual settings.

% \todo{ADD SOME RESULTS NUMBERS / RANKINGS HERE}.

% SemEval-2026 Task 11 aims to investigate how language models can acquire content-independent multilingual reasoning mechanisms, thereby mitigating content biases that affect their logical reasoning capabilities across languages. 

% We propose a shared task on multilingual syllogistic reasoning to improve our understanding of how to disentangle content from formal reasoning in LLMs. In this task, participants will be presented with syllogistic arguments in different languages that can be aligned (i.e. plausible) or misaligned (i.e. implausible) with world knowledge. The goal is to build models that can assess the formal validity of the arguments, regardless of their plausibility

% \begin{itemize}
%     \item What is the task about and why is it important? Be sure to mention the language(s) covered and cite the task overview paper. ~1 paragraph
%     \item What is the main strategy your system uses? ~1 paragraph
%     \item What did you discover by participating in this task? Key quantitative and qualitative results, such as how you ranked relative to other teams and what your system struggles with. ~1 paragraph
%     \item Have you released your code? Give a URL
% \end{itemize}

\section{Background}
\paragraph{Categorical Syllogisms}
Categorical syllogisms are a compact form of deductive reasoning consisting of two premises and a conclusion \citep{prior1962formal,ramsey2009problem,priest2008introduction}. Their validity is entirely determined by structural configuration, making them a natural benchmark for evaluating whether models follow logical form rather than surface cues \cite{wu-etal-2023-hence,ozeki2024exploring}. In practice, the core challenge lies in mapping natural language text onto the intended quantifiers, negations, and term relations, a process that is brittle under paraphrase and compounds across languages \cite{zong-lin-2024-categorical,cui-etal-2022-generalized-quantifiers}. 

\paragraph{Logical Structure and Terminology}
A categorical syllogism uses three terms. The subject of the conclusion is the minor term ($S$), the predicate of the conclusion is the major term ($P$), and the term that appears in both premises but not in the conclusion is the middle term ($M$). For a valid syllogism, the conclusion is always a claim about the relation between $S$ and $P$ \citep{eisape2024systematic}. In other words, $M$ is the shared handle that allows information to flow from one premises to the other and is then eliminated to produce a statement purely about $S$ and $P$.

\paragraph{Mood, Figure, and Validity}
Classical syllogistic theory encodes statements as four preposition types (Table~\ref{tab:logical-not}), with \textit{mood} defined as the ordered triple across major premise, minor premise, and conclusion, and \textit{figure} specifying the middle term's position, creating 256 possible forms of which only 24 are valid \citep{zong-lin-2024-categorical,eisape2024systematic, copi2014introduction}. Validity requires: middle term distribution in at least one premise, no valid conclusion from two negative premises, and exactly one negative premise for negative conclusions \citep{hurley2014concise}. Existential import enables subalternate moods like \textit{Barbari} and \textit{Darapti} \citep{parsons2014articulating},  making mood and figure as a compact notation and a complete decision procedure \citep{prior1962formal,ramsey2009problem,priest2008introduction}.

% Classical syllogistic theory encodes every statement as one of four preposition types, as summarized in Table~\ref{tab:logical-not}. The \textit{mood} of an argument is the ordered triple of these types across the major premise, minor premise, and conclusion, while the \textit{figure} specifies the position of $M$ as subject or predicate in each premise, yielding $256$ possible forms and only 24 are valid \citep{zong-lin-2024-categorical,eisape2024systematic, copi2014introduction}. Furthermore, validity is governed by distribution rules: the middle term be distributed in at least one premise, no valid conclusion follow fr`om two negative premises, and negative conclusion needs exactly one negative premise \citep{hurley2014concise}. Under Aristotelian logic, existential import further licenses subalternate moods such as \textit{Barbari} and \textit{Darapti} \citep{parsons2014articulating}, making mood and figure together both a compact notation and a complete decision procedure.

\paragraph{Trivial Validity}
Beyond the 24 structurally valid forms, some syllogistic arguments are formally valid for reasons that do not arise from the standard mood–figure interaction. These include \textit{petitio principii}, where the conclusion merely restates a premise \citep{walton2008begging}; immediate inferences such as valid conversion (restricted to $E$ and $I$ prepositions) and subalternation under existential import, where $A$ entails $I$ and $E$ entails $O$ \citep{hurley2014concise,parsons2014articulating}; and cases in which contradictory premises trigger vacuous validity via the principle of explosion (\textit{ex falso quodlibet}) \citep{priest2008introduction}.

\begin{figure*}[t]
  \includegraphics[width=1.01\linewidth, height=3.5cm]{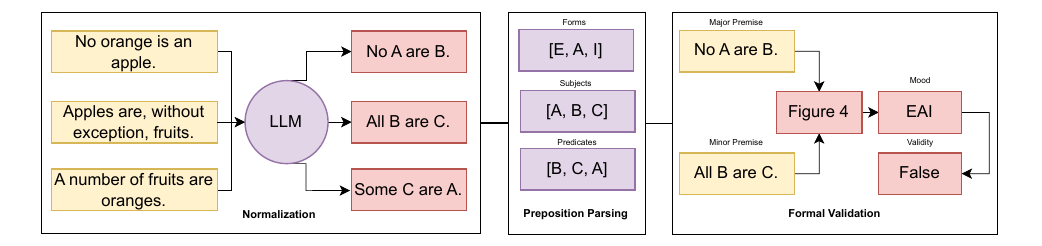}
  \caption {The flowchart illustrates the example step by step the flow of the proposed system. 
  % It starts with the process of \textbf{Normalization}, which transforms the premises into \textit{categorical syllogism notation}. For each preposition, the extraction of forms, subjects, and predicates is required for \textbf{Formal Validation}, which consists of premise categorization, figure setting, mood building, and validity checking.
  }
  \label{flowchart}
\end{figure*}

\section{System Overview}
% The general explanation of our proposed system is presented in the Figure \ref{flowchart}. \textbf{Normalization} is a process of converting each data point (two premises and one conclusion) into a categorical syllogism, \textbf{Preposition Parsing} is used to extract forms, subjects, and predicates, and the \textbf{Formal Validation} is used to determine the logical validity and its relevant premises.

% The detailed explanation of our proposed system is presented in the Figure \ref{flowchart}. Our system consists of three main steps, i.e., Normalization, Preposition Parsing, and Formal Validation. \textbf{Normalization} is a process of converting each data point (two premises and one conclusion) into a categorical syllogism, with each preposition being represented by terms such as A, E, I, or O. Meanwhile, \textbf{Preposition Parsing} is used to extract forms, subjects, and predicates that will then be used to infer the syllogistic \textit{mood} and \textit{figure}. Finally, the \textbf{Formal Validation} depends on the mood, figure, and its lookup validity table. 

\subsection{Normalization}
\subsubsection{Categorical Syllogism}
The categorical syllogistic normalization process is defined as a transformation function $f : \mathcal{N} \rightarrow \mathcal{C}$, where $\mathcal{N}$ denotes the space of natural-language syllogistic arguments and $\mathcal{C}$ denotes the space of categorical syllogistic representations. Given an input argument $a \in \mathcal{N}$, the model first identifies exactly three distinct semantic categories $\{T_1, T_2, T_3\}$ corresponding to the subject, predicate, and middle term of the syllogism. These terms are then abstracted into symbolic constants $\{A, B, C\}$ according to their order of first appearance in the argument.

For example, consider the natural-language argument: 
\begin{quote}
\textbf{Premise 1:} Some housecats enjoy chasing mice. \\
\textbf{Premise 2:} Any animal that enjoys chasing mice is a feline. \\
\textbf{Conclusion:} All cats are animals.
\label{example:1}
\end{quote}

The transformation function $f$ extracts three terms and maps them as:
\[
A:\text{animal}, \quad B:\text{feline}, \quad C:\text{cats}.
\]

The argument is then normalized into standard categorical form:
\[
\text{All } B \text{ are } A. \quad
\text{All } C \text{ are } A. \quad
\text{All } C \text{ are } B.
\]

\subsubsection{English Pivot Normalization}
As most LLMs generally perform better in English~\cite{guo-etal-2025-large}, to handle languages beyond English, all non-English syllogisms are first processed through a constrained translation procedure using an LLM~\footnote{The detailed prompts used for assessing logical validity and premise relevance are 
% not identical; see Appendix~\ref{sec:prompt_appendix} for the complete versions and differences.
shown in Appendix~\ref{sec:prompt_appendix}}. Instead of performing free-form translation, the model is instructed to extract the logical structure and translate only quantifiers and copular verbs into English. Furthermore, the original subject and predicate term is preserved in the the source language. This ensures structural standardization without introducing lexical drift that could alter term identity.

\subsection{Preposition Parsing}

After each data point is transformed into a canonical string of the form \texttt{P1. P2. Conclusion}, validity checking is extracted through a deterministic parsing procedure.

Each sentence $P_i$ is first matched against a constrained set of 
regular-expression patterns and mapped to one of the four categorical 
types $f_i \in \{A,E,I,O\}$, while extracting its subject and predicate 
terms $(s_i, p_i)$. Prior to matching, optional discourse markers 
(e.g., \textit{therefore}, \textit{thus}, \textit{hence}) and minor surface 
variations (e.g., \texttt{is}/\texttt{are}) are normalized to ensure 
form consistency. This produces a structured representation:
\[
\langle (f_1,s_1,p_1), (f_2,s_2,p_2), (f_3,s_3,p_3) \rangle.
\]

Let $S=s_3$ and $P=p_3$ denote the subject and predicate of the conclusion. 
The middle term $M$ is defined as the unique element in 
$(\{s_1,p_1\} \cap \{s_2,p_2\}) \setminus \{S,P\}$. 
The premise containing $P$ is identified as the major premise, 
and the premise containing $S$ as the minor premise. 
The figure is determined by the syntactic position (subject or predicate) 
of $M$ within the major and minor premises, yielding one of the four 
canonical configurations. The mood is computed as the ordered triple 
$f_{\text{major}} f_{\text{minor}} f_{\text{conclusion}}$. The resulting $(\text{mood}, \text{figure})$ pair serves as the input to the subsequent formal validation step.

\subsection{Formal Validation}
\label{sub: formal_validation}
\subsubsection{Logical Validity}

Given the parsed representation 
$\langle (f_1,s_1,p_1), (f_2,s_2,p_2), (f_3,s_3,p_3) \rangle$ 
and the inferred mood--figure pair $(m, \text{fig})$, 
logical validity is determined through a rule-based lookup procedure. 
For each figure $k \in \{1,2,3,4\}$, we define a predefined set of 
valid moods $\mathcal{V}_k$. The syllogism is classified as valid if

\[
\text{valid} = \mathbb{1}\{ m \in \mathcal{V}_{\text{fig}} \}.
\]

We additionally detect trivially valid cases 
(e.g., a premise identical to the conclusion or valid E/I converses) 
to avoid misclassifying degenerate arguments as invalid. 
The implementation details are provided in Appendix \ref{alg:parsing} and \ref{alg:parse}.

\subsubsection{Relevant Premises Identification}
For syllogisms classified as valid, the relevant-premise set $\mathcal{R} \subseteq \{1,2\}$ consists of the two premises that structurally connect $S$ and $P$ through the middle term $M$. Concretely, the major premise is the premise containing $P$, and the minor premise is the premise containing $S$. 
Thus, $\mathcal{R}$ contains exactly those two premise indices. Meanwhile, if the syllogism is classified as invalid, we define $\mathcal{R} = \emptyset$ by convention.

Using the previously introduced example in Section \ref{example:1}, the conclusion connects $S = C$ and $P = A$. The premise containing $P$ (``animal'') serves as the major premise, while the premise containing $S$ (``cats'') serves as the minor premise. These two premises form the structurally sufficient pair that links $S$ and $P$ 
through the middle term $M$. Therefore, $\mathcal{R} = \{1,2\}$, while any additional sentences, if present, are considered structurally irrelevant to the derivation.

\section{Experimental Setup}
\paragraph{Data Splits}
We use the official SemEval-2026 Task~11 data splits. The training set is used solely for prompt development and normalization strategy, while the development set is used for hyperparameter-free model comparison and ablation analysis. Final results are reported on the test set using the provided metric.

\paragraph{Normalization Model}
Premise normalization is performed using the Gemini 3 model, accessed via the official API.\footnote{We used the gemini-3-flash-preview} The model is used to transform raw inputs (english-only and multilingual) into canonical syllogistic form. 

\paragraph{Hyperparameter and Prompting}
All normalization prompts are fixed across splits. However, since the task includes english-only and multilingual premises, there are slight modifications to each task, following the experiment and ablation results. Furthermore, inference is performed with temperature $=0$ and seed $=0$ to ensure deterministic generation. No gradient-based fine-tuning is conducted.

\paragraph{Evaluation Metrics}
We report the official metrics defined in the shared task, including logical validity accuracy, Macro-averaged F1-Score for relevant premises, and combined score across english-only and multilingual tasks.

% \begin{itemize}
%     \item How data splits (train/dev/test) are used.
%     \item Key details about preprocessing, hyperparameter tuning, etc. that a reader would need to know to replicate your experiments. If space is limited, some of the details can go in an Appendix.
%     \item External tools/libraries used, preferably with version number and URL in a footnote.
%     \item Summarize the evaluation measures used in the task.
%     \item You do not need to devote much—if any—space to discussing the organization of your code or file formats.
% \end{itemize}

\section{Results}
In this section, we discuss the findings across all four subtasks (See Table \ref{tab:main-results} for detailed results) . The discussion is divided into two parts: validity, which focuses on Subtasks 1 (English-only) and 3 (Multilingual), and relevant premise, which focuses on Subtasks 2 (English-only) and 4 (Multilingual).

\subsection{Initial Experiment}
We conducted an initial experiment to identify the most effective normalization strategy. We compared Predicate-Argument (PA Notation), First-Order Logic (FOL Notation), and Syllogism (Categorical Syllogism) and evaluated them based on the given metric. Overall, as shown in Table~\ref{tab:normalization_comparison}, normalization substantially reduces content-effect bias compared to raw inputs. Among the approaches, standard syllogistic notation achieved the highest combined score, indicating an optimal balance between abstraction and model interpretability. In contrast, fully formal symbolic representations (e.g., $\forall$, $\exists$, $\rightarrow$, $\neg$) resulted in lower combined performance, as increased parsing complexity outweighed the gains from bias reduction.

\begin{table}[t]
  \centering
  \small
  \resizebox{\linewidth}{!}{
      \begin{tabular}{lccc}
        \hline
        \textbf{Normalization} & \textbf{Acc} & \textbf{Bias} & \textbf{Combined} \\
        \hline
        Raw Data & 92.10 & 10.63 & 26.66 \\
        PA Notation & 77.48 & 8.17 & 24.08 \\
        FOL Notation & 96.85 & 3.19 & 39.80 \\
        Syllogism Notation & \textbf{98.95} & \textbf{2.13} & \textbf{46.23} \\
        \hline
      \end{tabular}
  }
  \caption{Performance comparison across different normalization strategies on english-only data.}
  \label{tab:normalization_comparison}
\end{table}
\subsection{Validity Inference}

\subsubsection{English-only}
\label{sec:subtask1}
We have managed to achieve perfect accuracy with zero bias in English logical validity. It shows that the normalization step accurately maps natural language premises to their formal quantifier-term representations. Therefore, the parsing-based approach guarantee correctness by construction due to well-defined validity conditions. 

Conversely, the LLM-only setting decreased by approximately 2\% in accuracy, resulting in four mismatches: two false positives and two false negatives. The two false positives indicate plausibility bias, where the model accepts conclusions that appear semantically reasonable despite not being logically entailed by the premises. The two false negatives, by contrast, reflect difficulty in handling partitive quantifiers such as ``a number of,'' suggesting sensitivity to linguistic variation rather than plausibility alone. These errors are non-deterministic, as repeated runs may yield different misclassifications, further showing that pattern-matching-based reasoning is less reliable than deterministic symbolic resolution.

\subsubsection{Multilingual}
\label{sec:subtask3}
In line with the perfect accuracy and zero bias achieved in the English-only setting, the approach extends effectively to the multilingual condition through an English-pivot normalization strategy. Empirically, omitting the translation step (Norm + Parsing) leads to a performance drop of over 3\%, resulting in six structural mismatches. Error analysis indicates that, without translation, the cross-lingual variation in quantifier expression and term realization breaks the normalization step, producing unparseable structures that the deterministic rules cannot resolve.

Furthermore, the LLM-only baseline matches the accuracy of Norm + Parsing without translation, but with nearly double the bias score, and errors spanning six typologically diverse languages: Spanish, Swahili, Portuguese, Dutch, Bengali, and Russian. The false negatives mirror the English error patterns involving \textit{E}-type premises, while the single false positive occurs in Bengali, suggesting additional difficulty with non-Latin scripts. The increased bias relative to the English setting confirms that LLM reasoning degrades on non-English input, reinforcing the advantage of the translate-first strategy.

\begin{table}[!t]
    \centering
    \resizebox{\linewidth}{!}{
    \begin{tabular}{lcccc}
        \toprule
        \textbf{Method} & \textbf{Acc} & \textbf{F1 (Premise)} & \textbf{Bias} & \textbf{Combined} \\
        \toprule
        \multicolumn{5}{c}{Logical Validity (English)} \\
        \midrule
        LLM-only & 98.43 & -  & 2.13 & 45.74  \\
        Norm + Parsing & \textbf{100} & - & \textbf{0.0} & \textbf{100} \\
        \midrule
        \multicolumn{5}{c}{Relevance Premises (English)} \\
        \midrule
        LLM-only & 95.78 & \textbf{98.94} & 5.0 & 34.87 \\
        Norm + Parsing & \textbf{98.94} & 95.43 & \textbf{2.0} & \textbf{46.31} \\
        \midrule
        \multicolumn{5}{c}{Logical Validity (Multilingual)} \\
        \midrule
        LLM-only & 96.87 & - & 4.16 & 36.66 \\
        Norm + Parsing & 96.88 & - & 3.12 & 40.08  \\
        EPN + Norm + Parsing & \textbf{100} & - & \textbf{0.0} & \textbf{100 }\\
        \midrule
        \multicolumn{5}{c}{Relevance Premises (Multilingual)} \\
        \midrule
        LLM-only & 86.98 & 87.76 & 7.29 & 28.05  \\
        Norm + Parsing & 90.63 & 72.50 & 7.47 & 26.01 \\
        EPN + Norm + Parsing & \textbf{90.63} & \textbf{90.10} & \textbf{3.00} & \textbf{37.88}\\
        \bottomrule
    \end{tabular}
    }
    \caption{Comparison of different methods across subtasks. \textbf{LLM-only} denotes direct inference, \textbf{Norm + Parsing} is the deterministic parsing, and \textbf{EPN} refers to English Pivot Normalization.}
    \label{tab:main-results}
\end{table}

\subsection{Relevance Premises}
\subsubsection{English-only}
\label{sec:subtask2}
The deterministic method (Norm + Parsing) performs strongly, achieving 98.94 accuracy and a 95.43 F1-score in English relevance premise identification. Although the LLM-only attains a slightly higher premise-level F1-score, 98.94, it is more susceptible to content effects, which ultimately reduces its overall combined score. Compared to validity inference, relevant-premise identification is inherently more challenging, as it requires selecting the structurally necessary premise. For example, minor representational overlaps or redundancies between premises can lead to prediction mismatches See Appendix \ref{sec:subtask2_example} for more details.

Out of 190 instances, 16 mismatches were observed in premise prediction. The remaining discrepancies largely stem from overlapping universal statements. For example, in one case, where both premises express universal relations involving the same middle term, the system selects the structurally sufficient premise connecting the predicate term and omits an additional universal statement. This is logically compatible but not strictly required under the structural criterion. These mismatches reflect representational redundancy rather than fundamental reasoning errors.

\subsubsection{Multilingual}
\label{sec:subtask4}

The improvements observed in the English-only setting largely persist in the multilingual evaluation. The EPN+Norm+Parsing and Norm+Parsing approaches achieve higher validity accuracy (90.63) than the LLM-only baseline (86.98). In contrast, the structured approaches derive validity deterministically from the identified mood and figure. Norm+Parsing, by comparison, constructs a locally coherent structure that maps to a valid mood even when the original argument is invalid, due to inadvertent incorporation of distractor premises. For premise selection, EPN+Norm+Parsing attains a higher F1 score (90.10) than Norm+Parsing (72.50), primarily because Norm+Parsing suffers from rigid singular--plural distinctions (e.g., \textit{pianta/piante}, \textit{rosa/rose}) that cause normalization failures during regular-expression matching.

The LLM-only model reasons holistically over the full multi-sentence input and is frequently distracted by irrelevant premises, leading it to miss the logically active pair, while EPN+Norm+Parsing may fail when a distractor sentence is selected as a premise. The LLM-only baseline frequently selects semantically related distractors, whereas EPN-based methods introduce errors when the LLM itself selects a distractor sentence as a premise during the EPN step, directly outputting an incorrect canonical form before any downstream processing occurs. See Appendix~\ref{sec:subtask4_example} for more examples.

% \begin{itemize}
%     \item Main quantitative findings: How well did your system perform at the task according to official metrics? How does it rank in the competition?
%     \item Quantitative analysis: Ablations or other comparisons of different design decisions to better understand what works best. Indicate which data split is used for the analyses (e.g. in table captions). If you modify your system subsequent to the official submission, clearly indicate which results are from the modified system.
%     \item Error analysis: Look at some of your system predictions to get a feel for the kinds of mistakes it makes. If appropriate to the task, consider including a confusion matrix or other analysis of error subtypes—you may need to manually tag a small sample for this
% \end{itemize}

\subsection{Content-Effect Bias Reduction}
\begin{figure}[!t]
    \centering
    \includegraphics[width=\linewidth]{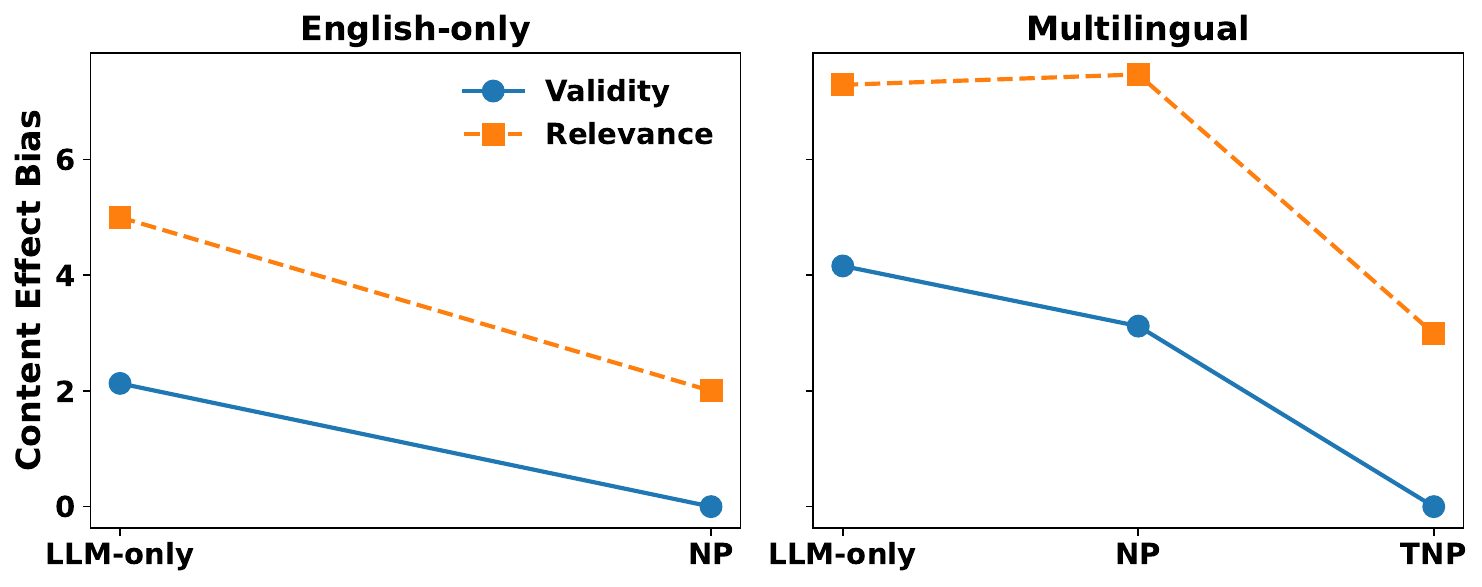}
    \caption{Content-effect reduction in English-only and Multilingual}
    \label{fig:radar_charts}
    \vspace{-0.8em}
\end{figure}

Across both validity inference and premise identification, our method substantially reduces content-effect bias compared to LLM-only baselines (Figure~\ref{fig:radar_charts}). In the English-only setting, Norm+Parsing eliminates validity bias entirely and reduces relevance bias from 5.0 to 2.0. A similar reduction trend is observed in multilingual evaluation: validity bias decreases from 4.16 (LLM-only) to 3.12 (Norm+Parsing) and to 0.0 under ENP+Norm+Parsing, while relevance bias drops markedly from 7.29 (LLM-only) and 7.47 (Norm+Parsing) to 2.99 with translation-based normalization. These consistent reductions highlight our hypothesis that formal syllogistic structure effectively mitigates content-effect biases across tasks and languages. By abstracting world-specific lexical through normalization and retaining only formal syllogistic structure, the deterministic method directly targets this source of interference.

\section{Conclusion}
By transforming syllogisms into canonical representations and applying deterministic parsing, our structural abstraction method achieves top-5 rankings across all SemEval-2026 Task 11 subtasks while substantially reducing content effects. This simple approach offers a compelling alternative to complex architectural modifications and opens promising avenues for developing scalable, interpretable reasoning techniques across languages.

% Our structural abstraction approach provides a simple yet effective solution for disentangling content from formal reasoning in LLMs through canonical syllogistic representation that preserves only logical structure, followed by deterministic structural parsing to determine validity. This method achieves competitive performance with a top-5 ranking across all SemEval-2026 Task 11 subtasks, demonstrating strong validity accuracy while substantially reducing content effects. Our approach offers a robust alternative to complex architectural modifications and suggests promising directions for future work in developing scalable, interpretable methods for mitigating reasoning biases across diverse linguistic contexts.
 
% SemEval-2026 Task 11 aims to investigate how language models can acquire content-independent multilingual reasoning mechanisms, thereby mitigating content biases that affect their logical reasoning capabilities across languages. 

% \begin{itemize}
%     \item A few summary sentences about your system, results, and ideas for future work.
% \end{itemize}

% \section*{Limitations}

% This document does not cover the content requirements for ACL or any
% other specific venue.  Check the author instructions for
% information on
% maximum page lengths, the required ``Limitations'' section,
% and so on.

\section*{Acknowledgments}
We sincerely thank Onno Kampman from SEACrowd for generously providing Google Gemini credits, which made the execution of our experiments possible.

\section*{Limitations}
This paper evaluates only one commercial model, Gemini-3-Flash, and does not explore other commercial or open-source models. In addition, we use a fully deterministic decoding strategy (greedy decoding with temperature set to 0 and a fixed seed). We do not conduct experiments across multiple random seeds or sampling settings, and therefore do not examine performance variability or leverage the potential diversity of large language models.

\section*{Ethical Considerations}
We acknowledge that our research utilized AI tools for writing, rewriting, and generating code. Although these tools offer significant advantages in terms of efficiency and productivity, their use raises important ethical considerations. We recognize the potential for bias and errors inherent in AI-generated content and have taken steps to mitigate these risks through rigorous human review and validation. Furthermore, we are mindful of the potential impact on the broader software development community, particularly regarding job displacement and the need for upskilling. We believe that responsible AI integration should prioritize transparency, accountability, and the empowerment of human developers, ensuring that these tools augment rather than replace human expertise. This research aims to contribute to the ongoing dialogue on ethical AI development and usage, advocating for a future where AI tools are harnessed responsibly to enhance human creativity and innovation in the field of software engineering.

\bibliography{custom}

\appendix

\section{Leaderboard Comparison}
\label{sec:leaderboard_comparison}
Below we present the top-5 comparison results for each subtask: Table~\ref{tab:leaderboard-st1} corresponds to Subtask 1, Table~\ref{tab:leaderboard-st2} to Subtask 2, Table~\ref{tab:leaderboard-st3} to Subtask 3, and Table~\ref{tab:leaderboard-st4} to Subtask 4. Our team is registered as \textbf{itlc\_team} on Codabench; for simplicity, we refer to it as \textbf{ITLC}. For Subtask 2 (Table~\ref{tab:leaderboard-st2}), our submission appears under the name \textbf{joanitolopo}, due to changes in the test data made by the organizers during the evaluation phase. In addition, the scores on the leaderboard differ from those reported in the current paper, particularly for Subtask 4, since a unified prompt was used across all subtasks in this paper. However, the overall ranking remains the same.

\begin{table}[!ht]
    \centering
    \resizebox{\linewidth}{!}{
    \begin{tabular}{lcccc}
        \toprule
        \textbf{\#} & \textbf{Team Name} & \textbf{Accuracy} & \textbf{Content Effect} & \textbf{Combined} \\
        \midrule
        1 & rongchuan & 100 & 0.0 & 100 \\
        ... & ... & ... & ... & ... \\
        \cellcolor{oursgray} 9 & \cellcolor{oursgray} ITLC & \cellcolor{oursgray} 100 & \cellcolor{oursgray} 0.0 & \cellcolor{oursgray} 100 \\
        ... & ... & ... & ... & ... \\
        12 & ewelinaksiez  & 97.91 & 0.02 & 95.81  \\
        13 & chisnguyen & 99.48 & 1.06 & 57.68  \\
        14 & vinaybabu & 99.48 & 1.06 & 57.68  \\
        \bottomrule
    \end{tabular}
    }
    \caption{Leaderboard comparison of Subtask 1}
    \label{tab:leaderboard-st1}
\end{table}

\begin{table}[!ht]
    \centering
    \resizebox{\linewidth}{!}{
    \begin{tabular}{lccccc}
        \toprule
        \textbf{\#} & \textbf{Team Name} &  \textbf{Accuracy} & \textbf{F1} & \textbf{Bias} & \textbf{Combined} \\
        \midrule
        1 & Habib\_TAZ & 100 & 100  & 0.0 & 100  \\
        2 & YNU-NLP & 100 & 100  & 0.0 & 100  \\
        3 & PA & 100 & 98.95  & 0.0 & 99.47  \\
        4 & junhaofu & 96.84 & 94.21  & 1.13 & 54.43  \\
        5 & butasrafael & 96.88 & 95.83  & 1.17 & 54.24  \\
        \cellcolor{oursgray}6 & \cellcolor{oursgray} ITLC & \cellcolor{oursgray} 98.94 & \cellcolor{oursgray} 95.43 & \cellcolor{oursgray} 2.0 & \cellcolor{oursgray} 46.31 \\
        \bottomrule
    \end{tabular}
    }
    \caption{Leaderboard comparison of Subtask 2}
    \label{tab:leaderboard-st2}
\end{table}

\begin{table}[!ht]
    \centering
    \resizebox{\linewidth}{!}{
    \begin{tabular}{lcccc}
        \toprule
        \textbf{\#} & \textbf{Team Name} & \textbf{Accuracy} & \textbf{Content Effect} & \textbf{Combined} \\
        \midrule
        1 & Habib\_TAZ & 100 & 0.0 & 100 \\
        2 & PA & 100 & 0.0 & 100 \\
        \cellcolor{oursgray}3 & \cellcolor{oursgray} ITLC & \cellcolor{oursgray} 100 & \cellcolor{oursgray} 0.0 & \cellcolor{oursgray} 100 \\
        4 & junhaofu  & 95.83 & 0.17 & 82.59  \\
        5 & rongchuan & 96.35 & 1.0 & 56.97  \\
        \bottomrule
    \end{tabular}
    }
    \caption{Leaderboard comparison of Subtask 3}
    \label{tab:leaderboard-st3}
\end{table}

\begin{table}[!ht]
    \centering
    \resizebox{\linewidth}{!}{
    \begin{tabular}{lccccc}
        \toprule
        \textbf{\#} & \textbf{Team Name} &  \textbf{Accuracy} & \textbf{F1} & \textbf{Bias} & \textbf{Combined} \\
        \midrule
        1 & PA & 100 & 94.36  & 0.0 & 97.18  \\
        2 & YNU-NLP & 90.1 & 89.58  & 1.26 & 49.51  \\
        3 & sungbin\_kai & 86.46 & 86.8  & 1.2 & 48.48  \\
        4 & ufal\_cuni & 84.9 & 83.42  & 1.37 & 45.2  \\
        \cellcolor{oursgray}5 & \cellcolor{oursgray} ITLC & \cellcolor{oursgray} 91.15 & \cellcolor{oursgray} 98.31 & \cellcolor{oursgray} 2.19 & \cellcolor{oursgray} 43.83 \\
        \bottomrule
    \end{tabular}
    }
    \caption{Leaderboard comparison of Subtask 4}
    \label{tab:leaderboard-st4}
\end{table}

\section{Parsing algorithm}
\label{alg:parsing}

\begin{algorithm}[ht]
\caption{Parse one preposition into (form, subj, pred)}
\label{alg:match-aeio}
\begin{algorithmic}[0]
\Function{MatchAEIO}{$p$}
  \State $p \gets$ lowercase($p$); trim($p$); 
  \State replace `` is '' with `` are ''
  \State remove leading connector in \{\textit{therefore, thus, hence, so}\} 
  \If{$p$ matches ``all $X$ are $Y$''} \State \Return $(A,X,Y)$ \EndIf
  \If{$p$ matches ``no $X$ are $Y$''} \State \Return $(E,X,Y)$ \EndIf
  \If{$p$ matches ``some $X$ are not $Y$''} \State \Return $(O,X,Y)$ \EndIf
  \If{$p$ matches ``some $X$ are $Y$''} \State \Return $(I,X,Y)$ \EndIf
  \State \Return fail
\EndFunction
\end{algorithmic}
\end{algorithm}

\section{Lookup Table}
\label{alg:parse}
\begin{algorithm}[t]
\caption{Infer mood/figure and validate by lookup}
\label{alg:validate-syllogism}
\footnotesize
\begin{algorithmic} % no [1] -> no line numbers
\Function{ValidateSyllogism}{$x$}
  \State $parts \gets$ \Call{SplitNonEmpty}{$x$,"."}
  \If{$|parts|\neq 3$} \State \Return $(\emptyset,0,\mathbf{false})$ \EndIf
  \State $(p_1,p_2,c)\gets(parts[1],parts[2],parts[3])$
  \State $(f_1,s_1,r_1,ok_1)\gets$\Call{MatchAEIO}{$p_1$}
  \State $(f_2,s_2,r_2,ok_2)\gets$\Call{MatchAEIO}{$p_2$}
  \State $(f_3,s_3,r_3,ok_3)\gets$\Call{MatchAEIO}{$c$}
  \If{\textbf{not} $(ok_1\land ok_2\land ok_3)$} \State \Return $(\emptyset,0,\mathbf{false})$ \EndIf

  \State $S\gets s_3,\ P\gets r_3,\ U_1\gets\{s_1,r_1\},\ U_2\gets\{s_2,r_2\}$
  \If{$|U_1\cup U_2\cup\{S,P\}|\neq 3$} \State \Return $(\emptyset,0,\mathbf{false})$ \EndIf

  \State $Mset\gets (U_1\cap U_2)\setminus\{S,P\}$
  \If{$|Mset|\neq 1$} \State \Return $(\emptyset,0,\mathbf{false})$ \EndIf
  \State $M\gets$\Call{Only}{$Mset$}

  \If{$P\in U_1$} \State $maj\gets 1$; $min\gets 2$
  \ElsIf{$P\in U_2$} \State $maj\gets 2$; $min\gets 1$
  \Else \State \Return $(\emptyset,0,\mathbf{false})$
  \EndIf

  \State $a\gets (s_{maj}=M)$; $b\gets (s_{min}=M)$
  \If{$a\land \neg b$} \State $figure\gets 1$
  \ElsIf{$\neg a\land \neg b$} \State $figure\gets 2$
  \ElsIf{$a\land b$} \State $figure\gets 3$
  \Else \State $figure\gets 4$
  \EndIf

  \State $mood\gets(f_{maj},f_{min},f_3)$
  \State \Return $(mood,figure,\;mood\in VALID[figure])$
\EndFunction
\end{algorithmic}
\end{algorithm}

\begin{table}[H]
\caption{Lookup table for valid moods by figure}
\label{tab:valid-moods}
\centering
\small
\begin{tabular}{@{}ll@{}}
\toprule
Figure & Valid moods (major--minor--conclusion) \\
\midrule
1 & AAA, EAE, AII, EIO, AAI, EAO \\
2 & EAE, AEE, EIO, AOO, EAO, AEO \\
3 & AAI, IAI, AII, EAO, OAO, EIO \\
4 & AAI, AEE, IAI, EAO, EIO, AEO \\
\bottomrule
\end{tabular}
\end{table}

% Requires: \usepackage{booktabs}

\begin{table*}[t]
\centering
\caption{Four types of categorical sentences and their translations into predicate-logic and set-theoretic notation.}
\label{tab:logical-not}
\small
\begin{tabular}{@{}lllll@{}}
\toprule
Type & Sentence Pattern & Predicate Logic & Set Theory & Description \\
\midrule
$A$ (all) & All $S$ are $P$ & $\forall x\,(S(x)\rightarrow P(x))$ & $S\subseteq P$ & Universal Affirmative \\
$E$ (no) & No $S$ are $P$ & $\forall x\,(S(x)\rightarrow \neg P(x))$ & $S\cap P=\emptyset$ & Universal Negative \\
$I$ (some) & Some $S$ are $P$ & $\exists x\,(S(x)\wedge P(x))$ & $S\cap P\neq\emptyset$ & Particular Affirmative \\
$O$ (some-not) & Some $S$ are not $P$ & $\exists x\,(S(x)\wedge \neg P(x))$ & $S\setminus P\neq\emptyset$ & Particular Negative \\
\bottomrule
\end{tabular}
\end{table*}

\section{Example Appendix}
\label{sec:example_appendix}

\subsection{Logical Validity Premises}
\subsubsection{English only}
\label{sec:subtask1_example}
As discussed in Section~\ref{sec:subtask1}, our normalization approach achieved 100\% accuracy due to the deterministic nature of preposition parsing. In contrast, Table~\ref{tab:plausibility-bias-example} illustrates a representative LLM-only error where the model predicts the argument as valid by relying on semantic plausibility rather than strict logical entailment. The conclusion ``some vehicles are bikes'' is true under general world knowledge, and this real-world truthfulness appears to bias the model toward accepting it. However, the conclusion is not derivable from the given premises: an E-type premise (``No bikes are cars'') paired with an A-type premise (``All bikes are vehicles'') does not license an existential affirmative conclusion about vehicles being bikes without assuming existential import. The deterministic parser correctly rejects this form because no valid mood-figure combination produces such a conclusion from the given premise types, regardless of whether the conclusion happens to be true in the real world. This distinction between semantic truth and logical validity is precisely where LLM-based inference breaks down, as the model conflates what is plausible with what is entailed.
\begin{table*}[ht]
\centering
\resizebox{\linewidth}{!}{
\renewcommand{\arraystretch}{1.4}
\begin{tabular}{cllll}
\toprule
\textbf{Idx} & \textbf{Content} & \textbf{Interpretation} & \textbf{Role} & \textbf{Inference Issue} \\
\midrule
0 & There are no bikes that can be called cars. & No bikes are cars. & premise & -- \\
1 & It is also true that every bike is a type of vehicle. & All bikes are vehicles. & premise & -- \\
\cellcolor{red!15}2 & This has led to the conclusion that a portion of vehicles are bikes. & Some vehicles are bikes. & \cellcolor{red!15}conclusion & \cellcolor{red!15}Semantically plausible, but not logically entailed. \\
\midrule
\multicolumn{5}{l}{GT validity: \textbf{False} \quad LLM-only prediction: \textbf{\textcolor{red}{True}}} \\
\bottomrule
\end{tabular}}
\caption{LLM-only (Subtask 1) exhibits plausibility bias by accepting a semantically plausible conclusion that is not logically entailed by the premises.}
\label{tab:plausibility-bias-example}
\end{table*}
\subsubsection{Multilingual}
\label{sec:subtask3_example}

The EPN + Norm + Parsing pipeline achieves perfect accuracy on multilingual validity as shown in Section \ref{sec:subtask3}, confirming that the deterministic approach generalizes fully across languages when EPN is applied. However, when relying on the LLM-only baseline, this guarantee breaks down. Table \ref{tab:error-negated-universal} illustrates a representative failure. The Spanish construction ``no se trata de que cada'' is a negated universal, which under standard categorical logic maps to a particular negative (O-type: ``Some $S$ are not $P$''). The LLM instead misinterprets the surface-level negation marker ``no'' as indicating a universal negative (E-type: ``No $S$ are $P$''), collapsing the distinction between sentential negation and quantifier negation. This misclassification propagates through the inference chain, causing the model to reject a valid conclusion. The error is particularly revealing because the underlying syllogistic structure is straightforward once the premises are correctly typed. The EPN-first pipeline avoids this entirely: translation preserves the negated-universal semantics in English (``it is not the case that every...''), and the normalization step deterministically maps this pattern to O-type before symbolic resolution is applied. This confirms that the bottleneck for multilingual validity is not logical reasoning itself, but quantifier interpretation across languages.

\begin{table*}[ht]
\centering
\resizebox{\linewidth}{!}{
\renewcommand{\arraystretch}{1.4}
\begin{tabular}{cp{5cm}p{4.5cm}p{3.5cm}lp{4.5cm}}
\toprule
\textbf{Idx} & \textbf{Content} & \textbf{Post EPN} & \textbf{Interpretation} & \textbf{Role} & \textbf{Error Note} \\
\midrule
0 & No se trata de que cada llave sea un objeto. & It is not the case that every key is an object. & Some keys are not objects. (O) & premise & -- \\
1 & Todo lo que sea una llave inglesa es una herramienta. & Everything that is a wrench is a tool. & All wrenches are tools. (A) & premise & -- \\
\cellcolor{red!15}2 & \cellcolor{red!15}Por tanto, se puede concluir que algunas herramientas no son objetos. & \cellcolor{red!15}Therefore, some tools are not objects. & \cellcolor{red!15}Some tools are not objects. (O) & \cellcolor{red!15}conclusion & \cellcolor{red!15}LLM fails to convert the negated universal (``no se trata de que cada'') into O-type; treats it as E-type, breaking the inference chain. \\
\midrule
\multicolumn{6}{l}{GT validity: \textbf{True} \quad LLM-only prediction: \textbf{\textcolor{red}{False}} \quad Language: \textbf{Spanish}} \\
\bottomrule
\end{tabular}}
\caption{LLM-only (Subtask 3) fails to resolve a negated universal quantifier in Spanish, misinterpreting ``no se trata de que cada'' as a universal negative rather than the intended particular negative.}
\label{tab:error-negated-universal}
\end{table*}

Table \ref{tab:error-term-collapse} illustrates the primary failure mode of the Norm + Parsing pipeline when applied without prior translation. The original English syllogism contains three distinct terms (``dog,'' ``poodle,'' ``canine'') and is straightforwardly valid. However, French lacks a lexical distinction between ``dog'' and ``canine,'' collapsing both into ``chien.'' This translation-induced term collapse reduces the second premise from a meaningful categorical statement (``All dogs are canines'') to a tautology (``All chiens are chiens''), producing a degenerate two-term structure that the parser assigns Figure 0 with undefined mood. The deterministic rules therefore reject the argument despite the original being logically sound. This pattern recurs across five of the six errors in the Norm + Parsing setting, where cross-lingual lexical gaps or synonym merging similarly destroy the term structure required for symbolic resolution. The EPN + Norm + Parsing pipeline avoids this entirely by operating on the English source, where the three-term distinction is preserved by construction.

\begin{table*}[ht]
\centering
\resizebox{\linewidth}{!}{
\renewcommand{\arraystretch}{1.4}
\begin{tabular}{cp{4.5cm}p{5cm}p{3.5cm}lp{4.5cm}}
\toprule
\textbf{Idx} & \textbf{Original (English)} & \textbf{Content (French)} & \textbf{Interpretation} & \textbf{Role} & \textbf{Error Note} \\
\midrule
0 & It is not true that every dog is a poodle. & Ce n'est pas vrai que tous les chiens sont des caniches. & Some dogs are not poodles. (O) & premise & -- \\
\cellcolor{red!15}1 & \cellcolor{red!15}Every creature that is a dog is a \textbf{canine}. & \cellcolor{red!15}Toute cr\'{e}ature qui est un \textbf{chien} est un \textbf{chien}. & \cellcolor{red!15}All A are A. (tautology) & \cellcolor{red!15}premise & \cellcolor{red!15}French collapses ``canine'' into ``chien'' (dog), destroying the three-term structure. \\
2 & Some \textbf{canines} are not poodles. & Certains \textbf{chiens} ne sont pas des caniches. & Some dogs are not poodles. (O) & conclusion & -- \\
\midrule
\multicolumn{6}{l}{GT validity: \textbf{True} \quad Norm + Parsing prediction: \textbf{\textcolor{red}{False}} \quad Language: \textbf{French}} \\
\bottomrule
\end{tabular}}
\caption{Norm + Parsing without translation (Subtask 3) fails on a French syllogism where the translation collapses ``canine'' and ``dog'' into the same French word ``chien,'' reducing a valid three-term syllogism to a degenerate two-term structure.}
\label{tab:error-term-collapse}
\end{table*}

\subsection{Relevance Premises}
\subsubsection{English-only}
\label{sec:subtask2_example}
Table \ref{tab:minor-overlap-example} illustrates the example of minor representational overlaps or redundancies between premises. Table \ref{tab:llm-only-mismatch-example} shows that LLM failed to select any relevant premises, while Table \ref{tab:false-positive-selection} shows false positive selection example by LLM. 

\begin{table*}[ht]
\centering
\resizebox{\linewidth}{!}{
\renewcommand{\arraystretch}{1.3}
\begin{tabular}{clll}
\toprule
\textbf{Idx} & \textbf{Premise} & \textbf{Role} & \textbf{Prediction} \\
\midrule
\cellcolor{green!15}0 & There are no circles that are also three-sided figures. & Relevant & \cellcolor{green!15}Selected \\
\cellcolor{wrong}1 & Some isosceles triangles are three-sided figures. & Distractor & \cellcolor{wrong}Wrongly selected \\
2 & All scalene triangles are three-sided figures. & Distractor & -- \\
3 & Every equilateral triangle is a three-sided figure. & Distractor & -- \\
\cellcolor{green!15}4 & All figures that are triangles are three-sided figures. & Relevant & \cellcolor{green!15}Missed \\
5 & Circles are not triangles. & Distractor & -- \\
\midrule
\multicolumn{4}{l}{\textbf{Gold Relevant Premises:} [0, 4] \quad \textbf{Predicted Relevant Premises:} [0, 1]} \\
\midrule
\multicolumn{4}{l}{\textbf{Observation:} Premise (1) shares the surface concept \textit{three-sided figures} with the gold premises,} \\
\multicolumn{4}{l}{creating a minor representational overlap that attracts selection} \\
\bottomrule
\end{tabular}}
\caption{Example of minor representational overlap in Subtask 4. The model selects premise (1) due to shared surface terminology (“three-sided figures”), although premise (4) is required for the correct logical structure.}
\label{tab:minor-overlap-example}
\end{table*}

\begin{table*}[ht]
\centering
\resizebox{\linewidth}{!}{
\renewcommand{\arraystretch}{1.3}
\begin{tabular}{clll}
\toprule
\textbf{Idx} & \textbf{Premise} & \textbf{Role} & \textbf{Prediction} \\
\midrule
0 & Some butterflies have colorful wings. & Distractor & -- \\
1 & It is true that all beetles have six legs. & Distractor & -- \\
\cellcolor{green!15}2 & There are no webs spun by grasshoppers. & Relevant & \cellcolor{green!15}Missed \\
\cellcolor{green!15}3 & Spiders are never ants. & Relevant & \cellcolor{green!15}Missed \\
4 & Every single bee pollinates flowers. & Distractor & -- \\
5 & A few ants are, in fact, insects. & Distractor & -- \\
6 & There exist insects that are not spiders. & Distractor & -- \\
\midrule
\multicolumn{4}{l}{\textbf{Gold Relevant Premises:} [2, 3] \quad \textbf{Predicted Relevant Premises:} []} 
\\
\midrule
\multicolumn{4}{l}{\textbf{Observation:} Instead of tracking the negation relations required for the valid inference, LLM overlooks the} \\
\multicolumn{4}{l}{necessary premises entirely, leading to an incorrect validity prediction.} \\
\bottomrule
\end{tabular}}
\caption{Example of LLM-only relevant premises mismatch. Although premises (2) and (3) are required to establish the contradiction structure, the model fails to select any relevant premises and predicts an incorrect validity label.}
\label{tab:llm-only-mismatch-example}
\end{table*}

\begin{table*}[ht]
\centering
\resizebox{\linewidth}{!}{
\renewcommand{\arraystretch}{1.3}
\begin{tabular}{clll}
\toprule
\textbf{Idx} & \textbf{Premise} & \textbf{Role} & \textbf{Prediction} \\
\midrule
0 & Some hounds are actually birds. & Distractor & -- \\
1 & Any retriever is a type of fish. & Distractor & -- \\
\cellcolor{red!15}2 & Anything that is a poodle is also a canine. & Distractor & \cellcolor{red!15}Wrongly selected \\
3 & There are no animals that bark. & Distractor & -- \\
4 & It is a fact that all collies have scales. & Distractor & -- \\
5 & Every single puppy is a kitten. & Distractor & -- \\
6 & The entire set of poodles is contained within the set of dogs. & Distractor & -- \\
\cellcolor{red!15}7 & A portion of dogs are not canines. & Distractor & \cellcolor{red!15}Wrongly selected \\
\midrule
\multicolumn{4}{l}{\textbf{Gold Relevant Premises:} [] \quad \textbf{Predicted Relevant Premises:} [2, 7]} \\
\midrule
\multicolumn{4}{l}{\textbf{Observation:} The model selects premises (2) and (7) due to lexical and conceptual overlap between \textit{poodles, dogs,} } \\
\multicolumn{4}{l}{and \textit{canines}, forming an apparent contradiction. However, these premises are structurally irrelevant to the target} \\
\multicolumn{4}{l}{ conclusion, resulting in a false-positive relevance attribution and incorrect validity judgment.} \\
\bottomrule
\end{tabular}}
\caption{Example of false-positive premise selection. Although no premise is structurally required for the invalid syllogism, the model selects (2) and (7) due to surface-level semantic overlap, leading to an incorrect validity prediction.}
\label{tab:false-positive-selection}
\end{table*}

\subsubsection{Multilingual}
\label{sec:subtask4_example}

\begin{table*}[ht]
\centering
\resizebox{\linewidth}{!}{
\renewcommand{\arraystretch}{1.4}
\begin{tabular}{clll}
\toprule
\textbf{Idx} & \textbf{Original (IT)} & \textbf{English} & \textbf{Role} \\
\midrule
\cellcolor{selected}0 & Tutto ciò che è un uccello depone le uova & Anything that is a bird lays eggs & \cellcolor{selected}GT premise \\
\cellcolor{distractor}1 & Ogni singola anatra è un uccello & Every single duck is a bird & Distractor \\
\cellcolor{distractor}2 & Alcuni uccelli sono creature che costruiscono nidi & Some birds are creatures that build nests & Distractor \\
\cellcolor{distractor}3 & Qualsiasi struzzo è un uccello che depone le uova & Any ostrich is a bird that lays eggs & Distractor \\
\cellcolor{distractor}4 & Non esistono pinguini che non depongono le uova & There are no penguins that do not lay eggs & Distractor \\
\cellcolor{selected}5 & Non esistono polli che non siano uccelli & There are no chickens that are not birds & \cellcolor{selected}GT premise \\
6 & È il caso che alcune galline depongano le uova & It is the case that some chickens lay eggs & Conclusion \\
\midrule
\multicolumn{4}{l}{GT premises: [0, 5] \quad LLM-only prediction: \textbf{[]} (predicted invalid)} \\
\bottomrule
\end{tabular}}
\caption{LLM-only (subtask 4) misses the active premise pair [0,5]; semantically plausible distractors about ostriches and penguins lead the model to predict invalid.}
\label{tab:llm-distractor}
\end{table*}

\begin{table*}[ht]
\centering
\resizebox{\linewidth}{!}{
\renewcommand{\arraystretch}{1.4}
\begin{tabular}{clll}
\toprule
\textbf{Idx} & \textbf{Original (ES)} & \textbf{English} & \textbf{Role} \\
\midrule
\cellcolor{distractor}0 & Cada brócoli es una verdura & Every single broccoli is a vegetable & Distractor \\
\cellcolor{distractor}1 & Es cierto que todas las frutas son comestibles & It is true that all fruits are edible & Distractor \\
\cellcolor{distractor}2 & Cualquier patata es una verdura & Any potato is a vegetable & Distractor \\
\cellcolor{distractor}3 & Algunas cosas comestibles son granos & Some edible things are grains & Distractor \\
\cellcolor{distractor}4 & Hay algunas raíces que son comestibles & There are a few roots that are edible & Distractor \\
\cellcolor{wrong}5 & Cada vegetal es comestible & Every single vegetable is edible & \cellcolor{wrong}Wrongly selected as P1 \\
\cellcolor{wrong}6 & Todo lo que es una zanahoria es una verdura & Anything that is a carrot is a vegetable & \cellcolor{wrong}Wrongly selected as P2 \\
7 & No hay zanahorias que no sean comestibles & There are no carrots that are not edible & Conclusion \\
\midrule
\multicolumn{4}{l}{Norm output: \textit{``All verdura are comestible. All zanahoria are verdura.Therefore, All zanahoria are comestible.''}} \\
\multicolumn{4}{l}{GT Validity: False \quad Norm+Parsing Prediction: \textbf{True}} \\
\bottomrule
\end{tabular}}
\caption{Norm+Parsing (subtask 4) selects sentences [5,6] forming a plausible carrot$\to$vegetable$\to$edible chain that maps to a valid mood, despite the overall argument being invalid.}
\label{tab:NP-plausible}
\end{table*}

\begin{table*}[ht]
\centering
\resizebox{\linewidth}{!}{
\renewcommand{\arraystretch}{1.4}
\begin{tabular}{clll}
\toprule
\textbf{Idx} & \textbf{Original (FR)} & \textbf{English} & \textbf{Role} \\
\midrule
\cellcolor{distractor}0 & Chaque chiot naît en sachant parler français & Every single puppy is born knowing how to speak French & Distractor \\
\cellcolor{distractor}1 & Aucun chien n'a jamais marché sur quatre pattes & No canine has ever walked on four legs & Distractor \\
\cellcolor{selected}2 & Il n'existe pas d'animaux qui soient aussi des chiens & There exist no animals that are also canines & \cellcolor{selected}GT premise \\
\cellcolor{distractor}3 & Il est indéniable que le ciel est vert et l'herbe bleue & The sky is green and the grass is blue & Distractor \\
\cellcolor{distractor}4 & Tous les chiens sont des robots déguisés & All dogs are actually disguised robots from space & Distractor \\
\cellcolor{wrong}5 & Certains animaux sont entièrement en verre & Some animals are made entirely of glass & \cellcolor{wrong}Wrongly selected as P1 \\
\cellcolor{selected}6 & L'ensemble des chiens fait partie des canines & The entire set of dogs is a part of the canines & \cellcolor{selected}GT premise \\
7 & Il n'y a pas d'animaux qui soient des chiens & There are no animals which are dogs & Conclusion \\
\midrule
\multicolumn{4}{l}{EPN output: \textit{``Some animaux are entièrement en verre. All chiens are canines. Therefore, No animaux are chiens.''}} \\
\multicolumn{4}{l}{GT premises: [2, 6] \quad EPN+Norm+Parsing Prediction: \textbf{[]} (predicted invalid)} \\
\bottomrule
\end{tabular}}
\caption{The EPN step (subtask 4) selects sentence [5] (a distractor) as P1 instead of the correct premise [2], producing a structure that the downstream classifier rejects as invalid.}
\label{tab:EPN-distractor}
\end{table*}

Table~\ref{tab:llm-distractor} illustrates a case where the LLM-only approach fails to identify the active premise pair due to semantically plausible distractors; Table~\ref{tab:NP-plausible} shows how Norm+Parsing constructs a locally coherent valid-looking structure from an invalid argument; and Table~\ref{tab:EPN-distractor} demonstrates a case where the EPN step itself selects a distractor as a premise, directly producing an incorrect canonical form before any downstream processing.

\section{Prompt Appendix}
\label{sec:prompt_appendix}

Figure~\ref{fig:llm_only_prompt} shows the LLM-only prompt used across all subtasks, Figure~\ref{fig:norm_prompt} shows the normalization prompt used across all subtasks, Figure~\ref{fig:subtask3_translation_prompt} shows the EPN prompt used in Subtask 3, Figure~\ref{fig:subtask4_translation_prompt} shows the EPN prompt used in Subtask 4, and Figure~\ref{fig:subtask4_translation_g_trans_prompt} shows its variant that incorporates the Google-translated sentence.

\begin{figure*}[!ht]
    \centering
    \begin{tcolorbox}[
        width=\linewidth,
        colback=white, % Background color
        colframe=black, % Border color
        sharp corners,
        boxrule=1pt, % Border thickness
        title={\textbf{LLM-only prompt}}, 
        fonttitle=\bfseries,
        coltitle=white,
        fontupper=\scriptsize,
    ]
\begin{verbatim}
You are a strict classical categorical logic evaluator.

You are given:
- An ID
- A syllogism containing multiple premises and one conclusion.

----------------------------------------
INSTRUCTIONS

1. All sentences before the final "Therefore" are premises.
   The sentence after "Therefore" is the conclusion.

2. Internally rewrite each statement into standard categorical form:
   - All X are Y (A)
   - No X are Y (E)
   - Some X are Y (I)
   - Some X are not Y (O)

   Handle paraphrases such as:
   - Every X is Y
   - Not a single X is Y
   - At least one X is not Y
   - Double negations

3. Determine whether the conclusion NECESSARILY follows
   from a subset of the premises under classical categorical logic.

4. If the argument is valid:
   - Identify the MINIMAL set of premises required to entail the conclusion.
   - Return their indexes (0-based).
   - Indexing is based on order of appearance in the text.
   - Do NOT include unused premises.

5. If the argument is invalid:
   - validity = false
   - relevant_premises = []

6. Do NOT explain reasoning.
   Output JSON ONLY.

STRICT REQUIREMENTS:
- Output must be valid JSON.
- No explanation.
- No markdown.
- No extra keys.
- Only "validity" and "relevant_premises".

----------------------------------------
OUTPUT FORMAT:

{{
  "validity": true or false,
  "relevant_premises": [int, int]
}}

----------------------------------------
SYLLOGISM:
{syllogism}
\end{verbatim}

\end{tcolorbox}
    \caption{LLM-only prompt for retrieve the validity and relevant premise directly}
    \label{fig:llm_only_prompt}
\end{figure*}

\begin{figure*}[!ht]
    \centering
    \begin{tcolorbox}[
        width=\linewidth,
        colback=white, % Background color
        colframe=black, % Border color
        sharp corners,
        boxrule=1pt, % Border thickness
        title={\textbf{Normalization prompt}}, 
        fonttitle=\bfseries,
        coltitle=white,
        fontupper=\scriptsize,
    ]
\begin{verbatim}
Transform a syllogistic argument into symbolic notation.

STEP 1 - IDENTIFY THE 3 TERMS:
- Find exactly 3 distinct categories/terms
- Middle term (M) = appears in BOTH premises, NEVER in conclusion
- Subject of conclusion (S) = subject in conclusion + appears in one premise
- Predicate of conclusion (P) = predicate in conclusion + appears in one premise

STEP 2 - MAP TERMS TO LETTERS:
- Assign A, B, C by order of first appearance in text

STEP 3 - CONVERT TO STANDARD FORM:
- "All X are Y" / "Every X is Y" / "X is subset of Y" / "Anything that is X is Y" → All X are Y (A)
- "No X are Y" / "Not a single X is Y" / "X cannot be Y" / "X is never Y" → No X are Y (E)
- "Some X are Y" / "A few X are Y" / "There exist X that are Y" / "Something that is X is Y" → Some X are Y (I)
- "Some X are not Y" / "Not all X are Y" / "At least one X is not Y" → Some X are not Y (O)
- "No X are not Y" / "There are no X that are not Y" → All X are Y (A) [double negative]

OUTPUT FORMAT (JSON only, no markdown):
{{
  "reasoning": "<free-form step-by-step analysis>",
  "mapped": "A:<term1>,B:<term2>,C:<term3>",
  "parsed": "<P1>. <P2>. <conclusion>"
}}

Transform:
{syllogism}
\end{verbatim}

\end{tcolorbox}
    \caption{Norm prompt for normalize sentences into standard categorical form}
    \label{fig:norm_prompt}
\end{figure*}

\begin{figure*}[!ht]
    \centering
    \begin{tcolorbox}[
        width=\linewidth,
        colback=white,
        colframe=black,
        sharp corners,
        boxrule=1pt,
        title={\textbf{EPN prompt for Validity Inference , Multilingual Setting (Subtask 3)}},
        fonttitle=\bfseries,
        coltitle=white,
    ]
    \begin{adjustbox}{max totalsize={\linewidth}{0.85\textheight}, center}
    \begin{minipage}{1.2\linewidth}
        \begin{Verbatim}[
        fontsize=\scriptsize,
        breaklines=false]
You are an expert translator for categorical syllogisms.

TASK
Given {syllogism}, extract its logical structure.
Translate quantifiers and verbs into English. Keep ONLY subject and predicate terms in the SOURCE language.

RULES:
- Output format per sentence: [Quantifier] [native_subj] are [native_pred]
- Quantifiers: All / No / Some / Some...not
- Never produce "All X are not Y".
- Preserve source subject as subject, source predicate as predicate. Do NOT swap.
- Remove rhetorical wrappers.

QUANTIFIER GUIDE:
- "every/all/any/anything that is X/each/whatever is" → All
- "no/none/never/there are no/not a single" → No
- "some/a few/certain/there are some/a portion" → Some
- "not all/not every" → Some...not
- "it is not the case that X are Y" with bare plurals → No.
- "anything that is X is Y" → All.

TERM DISTRIBUTION:
After extracting subj and pred for all 3 sentences, count how many sentences
each distinct native term appears in. Each term MUST appear in exactly 2 of 3 sentences.
Treat singular/plural variants as the same term (e.g. gari/magari).

If a term appears in 3 sentences and another appears in only 1:
CHECK: is the 1-count term a BROADER or NARROWER category of the 3-count term?
  - YES (e.g. maglietta/camicie = t-shirts/shirts): REPLACE one occurrence so both = 2.
  - NO (e.g. scrittori/individuo che scrive libri = writers/person who writes books,
    where one is a DEFINITION or DESCRIPTION, not a category): DO NOT replace.
    The 3-count term legitimately appears in all 3 sentences. Leave as-is.

If a term fills both slots in one sentence AND no other term covers its second sense:
tag it word[s] (specific) and word[g] (general).
Analyze verb context: relative clause ("thing that is X") = [s], main predicate = [g].
Propagate so each tagged term is in exactly 2 sentences.
Use the SAME base form for all tagged occurrences (pick one, use it everywhere).
Note: "poly: word → specific=[english], general=[english]".



REASONING FORMAT:
1. Extract subj/pred per sentence
2. Count term distribution
3. Fix distribution if needed (replace or tag)
4. Output

OUTPUT (JSON only, no markdown):
{{
  "detected_language": "<lang>",
  "reasoning": "<extract, count, fix, output>",
  "english": "P1. P2. Therefore, C."
}}

EXAMPLES:
Input: "All goyangi are dongmul. Some dongmul are not poyuryu. Therefore, some goyangi are not poyuryu."
reasoning: "Extract: P1 subj=goyangi pred=dongmul. P2 subj=dongmul pred=poyuryu. C subj=goyangi pred=poyuryu.
Count: goyangi=2, dongmul=2, poyuryu=2."
english: "All goyangi are dongmul. Some dongmul are not poyuryu. Therefore, some goyangi are not poyuryu."

Replacement needed (3-count and 1-count):
Input: "Todos los autos son máquinas. Hay carros que son carros. Algunos carros son máquinas."
reasoning: "Extract: P1 subj=autos pred=máquinas. P2 subj=carros pred=carros. C subj=carros pred=máquinas. 
Count: autos=1, máquinas=2, carros=3. Fix: replace P2 pred carros with autos. 
Count: autos=2, máquinas=2, carros=2. "
english: "All autos are máquinas. Some carros are autos. Therefore, Some carros are máquinas."

Replacement across sentences:
Input: "Nenhum gato é um pássaro. Todo gato é um animal doméstico. Pássaros não são gatos."
reasoning: "Extract: P1 subj=gato pred=pássaro. P2 subj=gato pred=animal doméstico. C subj=pássaro pred=gato. 
Count: gato=3, pássaro=2, animal doméstico=1. Fix: replace P1 subj gato with animal doméstico. Wait — that changes
the source subject. 
Instead: C says 'pássaros não são gatos' but gato=3, animal doméstico=1. They share a meaning. Replace C pred gato 
with animal doméstico. 
Count: gato=2, pássaro=2, animal doméstico=2. "
english: "No gato are pássaro. All gato are animal doméstico. Therefore, No pássaro are animal doméstico."

Tagging needed (same word both slots, no other term):
Input: "Hakuna samaki ni nyoka. Kila nyoka ni nyoka. Nyoka fulani si samaki."
reasoning: "Extract: P1 subj=samaki pred=nyoka. P2 subj=nyoka pred=nyoka. C subj=nyoka pred=samaki. 
Count: samaki=2, nyoka=4. No 1-count term to replace. P2 has same word both slots. Tag: P2 subj=nyoka[s] pred=nyoka[g]. 
Propagate: nyoka[s] in P2,C. nyoka[g] in P1,P2. samaki in P1,C. Each=2. . poly: nyoka → specific=snakes, general=reptiles."
english: "No samaki are nyoka[g]. All nyoka[s] are nyoka[g]. Some nyoka[s] are not samaki."

Polysemy with verb context:
Input: "Nicht alle Tiere sind Hunde. Jedes Geschöpf, das ein Tier ist, ist ein Tier. Daher sind einige Tiere keine Hunde."
reasoning: "Extract: P1 subj=Tiere pred=Hunde. P2 subj=Tier pred=Tier. C subj=Tiere pred=Hunde. Count: Tiere/Tier=4, Hunde=2. 
No 1-count term. P2 same word both slots. Tag using verb context: 'Geschöpf, das ein Tier ist' relative clause = [s], 'ist ein Tier' 
predicate = [g]. P2 subj=Tier[s] pred=Tier[g]. Propagate: Tier[s] in P1,P2. Tier[g] in P2,C. Hunde in P1,C. Each=2. . 
poly: Tier → specific=animals, general=creatures."
english: "Some Tiere[s] are not Hunde. All Tier[s] are Tier[g]. Therefore, Some Tiere[g] are not Hunde."

\end{Verbatim}
    \end{minipage}
    \end{adjustbox}
    \end{tcolorbox}
    \caption{EPN prompt for Subtask 3 for extract subject term}
    \label{fig:subtask3_translation_prompt}
\end{figure*}

\begin{figure*}[!ht]
    \centering
    \begin{tcolorbox}[
        width=\linewidth,
        colback=white, % Background color
        colframe=black, % Border color
        sharp corners,
        boxrule=1pt, % Border thickness
        title={\textbf{EPN prompt for Relevance Premises , Multilingual Setting (Subtask 4)}}, 
        fonttitle=\bfseries,
        coltitle=white,
        fontupper=\scriptsize,
    ]
\begin{verbatim}
You are an expert logical translator for categorical syllogisms.

TASK
Given {syllogism}, extract exactly:
- Two premises
- One conclusion

Preserve the original logical structure.
Translate ONLY quantifiers and copula into English.
Keep subject and predicate terms EXACTLY as in the source language.

STRICT RULES:

1. FORMAT
Each sentence must follow:
[Quantifier] [native_subject] are [native_predicate]

Allowed quantifiers:
All / No / Some / Some...not

Never output:
"All X are not Y"

2. DO NOT MODIFY LOGIC
- Do NOT swap subject and predicate.
- Do NOT replace terms.
- Do NOT merge synonyms.
- Do NOT normalize or repair the argument.
- Do NOT balance term distribution.
- Do NOT split polysemous words.
- Do NOT introduce [s] or [g] tags.

Extract the argument exactly as written.

3. TERM HANDLING
- Copy subject and predicate verbatim.
- Singular/plural variants count as the same term.
- Descriptive phrases remain descriptive phrases.
- Identity statements (X are X) remain unchanged.

4. TERM COUNT CHECK (Diagnostic Only)
After extraction:
- Count distinct terms.
- A classical syllogism should have exactly 3 distinct terms.
- If not, DO NOT repair.
- Simply report the count in reasoning.

5. SENTENCE SELECTION
If more than 3 sentences are present:
- Select the two premises and the conclusion that form the main argument.
- Ignore rhetorical commentary.

REASONING FORMAT:
1. Identify selected sentences (P1, P2, C)
2. Extract subject and predicate
3. Count distinct terms (no fixing)
4. Output final structured form

OUTPUT (JSON only, no markdown):
{
  "detected_language": "<lang>",
  "reasoning": "<selection + extraction + term_count>",
  "english": "P1. P2. Therefore, C."
}
\end{verbatim}

\end{tcolorbox}
    \caption{EPN prompt for Subtask 4 for extract and filter relevant premise and conclusion}
    \label{fig:subtask4_translation_prompt}
\end{figure*}

\begin{figure*}[!ht]
    \centering
    \begin{tcolorbox}[
        width=\linewidth,
        colback=white, % Background color
        colframe=black, % Border color
        sharp corners,
        boxrule=1pt, % Border thickness
        title={\textbf{EPN prompt with google translated sentence for Relevance Premises , Multilingual Setting (Subtask 4)}}, 
        fonttitle=\bfseries,
        coltitle=white,
        fontupper=\scriptsize,
    ]
\begin{verbatim}
You are an expert logical translator for categorical syllogisms.

TASK
Given Original syllogism in source language:{syllogism},
Google Translation reference:{google_translated_Sentence}
extract exactly:
- Two premises
- One conclusion

Preserve the original logical structure.
Translate ONLY quantifiers and copula into English.
Keep subject and predicate terms EXACTLY as in the source language.

STRICT RULES:

1. FORMAT
Each sentence must follow:
[Quantifier] [native_subject] are [native_predicate]

Allowed quantifiers:
All / No / Some / Some...not

Never output:
"All X are not Y"

2. DO NOT MODIFY LOGIC
- Do NOT swap subject and predicate.
- Do NOT replace terms.
- Do NOT merge synonyms.
- Do NOT normalize or repair the argument.
- Do NOT balance term distribution.
- Do NOT split polysemous words.
- Do NOT introduce [s] or [g] tags.

Extract the argument exactly as written.

3. GOOGLE TRANSLATE CHECK (Fidelity Only)
Mentally compare with Google Translate ONLY to:
- Verify quantifier accuracy
- Verify negation scope
- Verify copula meaning

4. TERM HANDLING
- Copy subject and predicate verbatim.
- Singular/plural variants count as the same term.
- Descriptive phrases remain descriptive phrases.
- Identity statements (X are X) remain unchanged.

5. TERM COUNT CHECK (Diagnostic Only)
After extraction:
- Count distinct terms.
- A classical syllogism should have exactly 3 distinct terms.
- If not, DO NOT repair.
- Simply report the count in reasoning.

6. SENTENCE SELECTION
If more than 3 sentences are present:
- Select the two premises and the conclusion that form the main argument.
- Ignore rhetorical commentary.

REASONING FORMAT:
1. Identify selected sentences (P1, P2, C)
2. Extract subject and predicate
3. Count distinct terms (no fixing)
4. Output final structured form

OUTPUT (JSON only, no markdown):
{
  "detected_language": "<lang>",
  "reasoning": "<selection + extraction + term_count>",
  "english": "P1. P2. Therefore, C."
}
\end{verbatim}

\end{tcolorbox}
    \caption{EPN prompt with google translated sentence for Subtask 4 for extract and filter relevant premise and conclusion}
    \label{fig:subtask4_translation_g_trans_prompt}
\end{figure*}

\section{Google Translation as a Double-Edged Sword in Multilingual Settings}
\begin{table*}[t]
  \centering
  \resizebox{\linewidth}{!}{
  \small
  \begin{tabular}{lcccc}
    \hline
    \textbf{Method} & \textbf{Acc} & \textbf{F1 (premise)} & \textbf{Bias} & \textbf{Combined} \\
    \hline
    Norm + Parsing & 90.63 & 72.50 & 7.47 & 26.01 \\
    Norm + Parsing using google translate sentence & 88.54 & 88.39 & 8.14 & 27.54 \\
    EPN + Norm + Parsing & \textbf{90.63} & \textbf{90.10} & 3.00 & \textbf{37.88} \\
    \quad + google translate example & 89.58 & 89.58 & 4.32 & 33.53 \\
    \hline
  \end{tabular}}
  \caption{Ablation study on the impact of incorporating Google-translated sentences in a multilingual setting}
  \label{tab:subtask4_ablation}
\end{table*}

\begin{table*}[ht]
\centering
\small
\setlength{\tabcolsep}{6pt}
\renewcommand{\arraystretch}{1.4}
\begin{tabular}{>{\bfseries}p{2.5cm} p{2.8cm} p{2.8cm} p{2.8cm} p{3.0cm}}
\toprule
 & \textbf{S1} & \textbf{S5} & \textbf{S7} & \textbf{Conclusion} \\
\midrule

Original (PT)
  & Alguns mamíferos não são cães
  & É certo que todo \textit{cachorro} é um mamífero
  & Todos os caninos são \textit{cães}
  & Todos os caninos são mamíferos \\

Google Trans.
  & Some mammals are not dogs
  & \cellcolor{selected}It is true that every \textcolor{colC}{\textbf{dog}} is a \textcolor{colB}{\textbf{mammal}}
  & \cellcolor{selected}All \textcolor{colA}{\textbf{canines}} are \textcolor{colC}{\textbf{dogs}}
  & \cellcolor{selected}All \textcolor{colA}{\textbf{canines}} are \textcolor{colB}{\textbf{mammals}} \\

Ground Truth
  & \cellcolor{selected}Some \textcolor{colB}{\textbf{mammals}} are not \textcolor{colC}{\textbf{dogs}}
  & Every single dog is a mammal
  & \cellcolor{selected}All \textcolor{colA}{\textbf{canines}} are \textcolor{colC}{\textbf{dogs}}
  & \cellcolor{selected}All \textcolor{colA}{\textbf{canines}} are \textcolor{colB}{\textbf{mammals}} \\

\bottomrule
\end{tabular}
\caption{Effect of lexical collapse in the Google-translated reference. Highlighted cells show selected premises per system. Terms are coloured consistently: \textcolor{colA}{\textbf{canines}}, \textcolor{colB}{\textbf{mammals}}, \textcolor{colC}{\textbf{dogs}}.}
\label{tab:translation-collapse}
\end{table*}

\begin{table*}[ht]
\centering
\small
\setlength{\tabcolsep}{6pt}
\renewcommand{\arraystretch}{1.4}
\begin{tabular}{>{\bfseries}p{2.5cm} p{3.2cm} p{3.2cm} p{3.2cm}}
\toprule
 & \textbf{P1} & \textbf{P2} & \textbf{Conclusion} \\
\midrule

Original (IT)
  & Tutto ciò che è una \textcolor{colC}{\textbf{rosa}} è anche una \textcolor{colB}{\textbf{pianta}}
  & Esistono \textcolor{colC}{\textbf{rose}} che sono \textcolor{colA}{\textbf{fiori}}
  & Alcuni \textcolor{colA}{\textbf{fiori}} sono \textcolor{colB}{\textbf{piante}} \\

Google Trans.
  & Everything that is a \textcolor{colC}{\textbf{rose}} is also a \textcolor{colB}{\textbf{plant}}
  & There are \textcolor{colC}{\textbf{roses}} that are \textcolor{colA}{\textbf{flowers}}
  & Some \textcolor{colA}{\textbf{flowers}} are \textcolor{colB}{\textbf{plants}} \\

\bottomrule
\end{tabular}
\caption{Italian syllogism with Google-translated reference. Terms are coloured consistently: \textcolor{colC}{\textbf{rose}}, \textcolor{colA}{\textbf{flowers}}, \textcolor{colB}{\textbf{plants}}.}
\label{tab:italian-syllogism}
\end{table*}

This section examines the impact of Google-translated sentences in the Relevance Premise (Multilingual) setting, which refer to subtask 4. We replace original sentences with their Google translations in Norm+Parsing, and provide translated sentences as references to the LLM in EPN+Norm+Parsing. Due to linguistic normalization, EPN often collapses synonyms and singular–plural distinctions into a single English form. This generally improves premise F1 but can harm validity accuracy, making translation a double-edged sword.

\subsection{Impact on Relevance Premise Selection}

As shown in Table~\ref{tab:italian-syllogism}, Google Translate collapses \textit{piante/pianta} and \textit{rosa/rose} into the same English terms (\textit{plant}, \textit{rose}). In Norm+Parsing, this enables successful matching using regular expressions and increases premise F1 from 72.50 to 88.39 (Table~\ref{tab:subtask4_ablation}).  

However, in EPN+Norm+Parsing, premise selection relies on LLM reasoning rather than regular expressions. Consequently, EPN slightly reduces F1 (90.10 to 89.58), as term collapse can distort the reasoning signal.

\subsection{Impact on Validity}

As illustrated in Table~\ref{tab:translation-collapse}, EPN collapses \textit{cachorro} and \textit{cães} into \textit{dog}, making the chain (\textit{canines} $\to$ \textit{dog} $\to$ \textit{mammal}) more transparent in English. This biases the LLM toward selecting an incorrect premise pair, producing a spurious validity form.  

Without EPN, the LLM explicitly treats \textit{cachorro} and \textit{cães} as distinct surface forms and avoids merging them. As a result, translation reduces validity accuracy and increases content bias: in Norm+Parsing, accuracy drops from 90.63 to 88.54 and bias rises from 7.47 to 8.14; in Translate+Norm+Parsing, accuracy decreases from 90.63 to 89.58 and bias increases from 2.99 to 4.32, as shown in Table~\ref{tab:subtask4_ablation}.

\end{document}